\documentclass[10pt,twocolumn,letterpaper]{article}

\usepackage[12]{cvpr} 

\usepackage{graphicx}
\usepackage{amsmath}
\usepackage{amssymb}
\usepackage{booktabs}

\usepackage[pagebackref,breaklinks,colorlinks]{hyperref}

\usepackage[capitalize]{cleveref}
\crefname{section}{Sec.}{Secs.}
\Crefname{section}{Section}{Sections}
\Crefname{table}{Table}{Tables}
\crefname{table}{Tab.}{Tabs.}

\usepackage[american]{babel}
\usepackage{microtype}
\newcommand*{\Scale}[2][4]{\scalebox{#1}{$#2$}}

\usepackage{booktabs} 
\usepackage{bm}

\usepackage{xcolor}
\usepackage{color, colortbl}
\usepackage{url}       
\usepackage{soul} 
\usepackage{multirow}
\usepackage{graphicx}

\usepackage{caption}

\usepackage[accsupp]{axessibility}

\usepackage{comment}

\newcommand{\myparagraph}[1]{\vspace{2pt}\noindent{\bf #1}}

\def\cvprPaperID{5161} 
\def\confName{CVPR}
\def\confYear{2022}

\author{
    Wenjia Xu$^{1,7,8}$ \hspace{0.4cm} Yongqin Xian$^2$ \hspace{0.4cm} Jiuniu Wang$^{5,7,8}$ \hspace{0.4cm} Bernt Schiele$^3$ \hspace{0.4cm} Zeynep Akata$^{3,4,6}$ \vspace{2mm} \\ 
    \noindent $^1$ Beijing University of Posts and Telecommunications $\quad$ $^2$ ETH Zurich $\quad$ \\ $^{3}$ Max Planck Institute for Informatics $\quad$ $^{4}$ University of T\"ubingen $\quad$ $^{5}$ City University of Hong Kong \\ $^{6}$ Max Planck Institute for Intelligent Systems \\
     $^{7}$ University of Chinese Academy of Sciences $\quad$ $^{8}$ Aerospace Information Research Institute, CAS
}

\begin{document}

\title{VGSE: Visually-Grounded Semantic Embeddings for Zero-Shot Learning}

\maketitle

\begin{abstract}
Human-annotated attributes serve as powerful semantic embeddings in zero-shot learning. However, their annotation process is labor-intensive and needs expert supervision. Current unsupervised semantic embeddings, i.e., word embeddings, enable knowledge transfer between classes. 
However, word embeddings do not always reflect visual similarities and result in inferior zero-shot performance.
We propose to discover semantic embeddings containing discriminative visual properties for zero-shot learning, without requiring any human annotation.
Our model visually divides a set of images from seen classes into clusters of local image regions according to their visual similarity, and further imposes their class discrimination and semantic relatedness.
To associate these clusters with previously unseen classes, we use external knowledge, e.g., word embeddings and propose a novel class relation discovery module.
Through quantitative and qualitative evaluation, we demonstrate that our model discovers semantic embeddings that model the visual properties of both seen and unseen classes. Furthermore, we demonstrate on three benchmarks that our visually-grounded semantic embeddings further improve performance over word embeddings across various ZSL models by a large margin. Code is available at https://github.com/wenjiaXu/VGSE

\end{abstract}


\section{Introduction}

Semantic embeddings aggregated for every class live in a vector space that associates different classes even when visual examples of these classes are not available. Therefore, 
they facilitate the knowledge transfer in zero-shot learning (ZSL)~\cite{xian2018zero,schonfeld2019generalized,ALE,liu2021goal} and are 
used as side-information in other computer vision tasks like fashion trend forecast~\cite{al2017fashion,yang2019interpretable,hsiao2017learning}, face recognition and manipulation~\cite{liu2015deep, lee2020maskgan,chen2018fsrnet}, and domain adaptation~\cite{chen2015deep,ishii2019zero}. 

\begin{figure}[t]
  \centering 
  \includegraphics[width=1\linewidth]{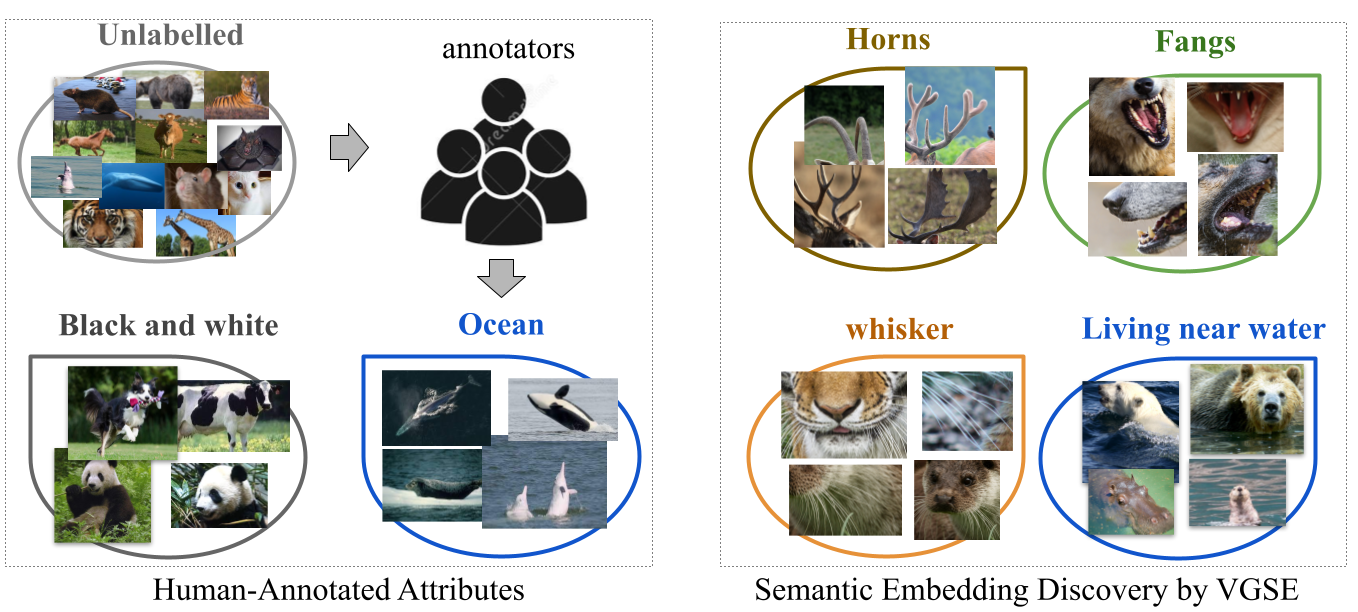}
    \caption{
    Human-annotated attributes (left) are labor-intensive to collect, and may neglect some local visual properties shared between classes. We propose to discover semantic embeddings via visually clustering image patches and predicting the class relations.
    }
  \label{fig:teaser}
\end{figure}

Human annotated attributes~\cite{25_SUNdataset,26_wah2011caltech,farhadi2009describing}, characteristic properties of objects annotated by human experts, are widely used as semantic embeddings~\cite{xian2019,xu2020attribute}. However, obtaining attributes is often a labor-intensive two-step process. First, domain experts carefully design an attribute vocabulary, e.g., color, shape, etc., and then human annotators indicate the presence or absence of an attribute in an image or a class~(as shown in Figure~\ref{fig:teaser}). The labeling effort devoted to human-annotated attributes hinders its applicability of performing zero-shot learning for more datasets in realistic settings~\cite{mall2021field}.

Previous works tackle this problem by using word embeddings for class names~\cite{w2v,glove}, or semantic embeddings from online encyclopedia articles~\cite{al2017automatic,zhu2018generative,qiao2016less}. Though they model the semantic relation between classes without using human annotation, some of these relations may not be visually detectable by machines, resulting in a poor performance in zero-shot learning. Similarly, discriminative visual cues may not all be represented in those semantic embeddings. 

To this end, we propose the Visually-Grounded Semantic Embedding~(VGSE) Network to discover semantic embeddings with minimal human supervision~(we only use category labels for seen class images). Our network explicitly explores visual clusters that relate image regions from different categories, which is useful for knowledge transfer between classes under zero-shot learning settings~(see our learnt clusters in Figure~\ref{fig:teaser}). 
To fully unearth the visual properties shared across different categories, our model discovers semantic embeddings by assigning image patches into various clusters according to their visual similarity.
Besides, we further impose class discrimination and semantic relatedness of the semantic embeddings, to benefit their ability in transferring knowledge between classes in ZSL.

To sum up, our work makes the following contributions.
(1) 
We propose a visually-grounded semantic embedding~(VGSE) network that learns visual clusters from seen classes, and automatically predicts the semantic embeddings for each category by building the relationship between seen and unseen classes given unsupervised external knowledge sources. 
(2) On three zero-shot learning benchmarks (i.e. AWA2, CUB, and SUN), our learned VGSE semantic embeddings consistently improve the performance of word embeddings over five SOTA methods.
(3) Through qualitative evaluation and user study, we demonstrate that our VGSE embeddings contain rich visual information like fine-grained attributes, and convey human-understandable semantics that facilitates knowledge transfer between classes.

\section{Related Work}

\myparagraph{Zero-shot Learning} aims to classify images from novel classes that do not appear during training. Existing ZSL methods usually assume that both the seen and unseen classes share a common semantic space, thus the key insight of performing ZSL is to transfer knowledge from seen classes to unseen classes. To assign the image to a semantic class embedding, many classical approaches learn a compatibility function to associate visual and semantic space
~\cite{ALE, CMT, xian2016latent, frome2013devise}.  Recent works mainly focus on synthesizing image features or classifier weights with a generative model~\cite{CLSWGAN,xian2019,schonfeld2019}, or training enhanced image features extractors with visual attention~\cite{ji2018stacked,SGMA} or local prototypes~\cite{xu2020attribute}. 

Semantic embeddings are crucial in relating different categories with shared characteristics, i.e., the semantic space. Despite their importance, semantic embeddings are relatively under-explored in zero-shot learning.
Human-annotated attributes~\cite{xian2018zero,25_SUNdataset,26_wah2011caltech,farhadi2009describing}, i.e., the properties of objects such as color and shape, are the most commonly used semantic embeddings in zero-shot learning. 
Though the attributes can be discriminative for each class, 
their annotation process is labor-intensive and require expert knowledge~\cite{song2018selective,yu2013designing,26_wah2011caltech}.
We propose to discover visual properties through patch-level clustering over image datasets, and predict semantic embeddings automatically, where no additional human annotation is required except for the class labels of seen class images. 

\begin{figure*}[tb]
  \centering 
  \includegraphics[width=0.95\linewidth]{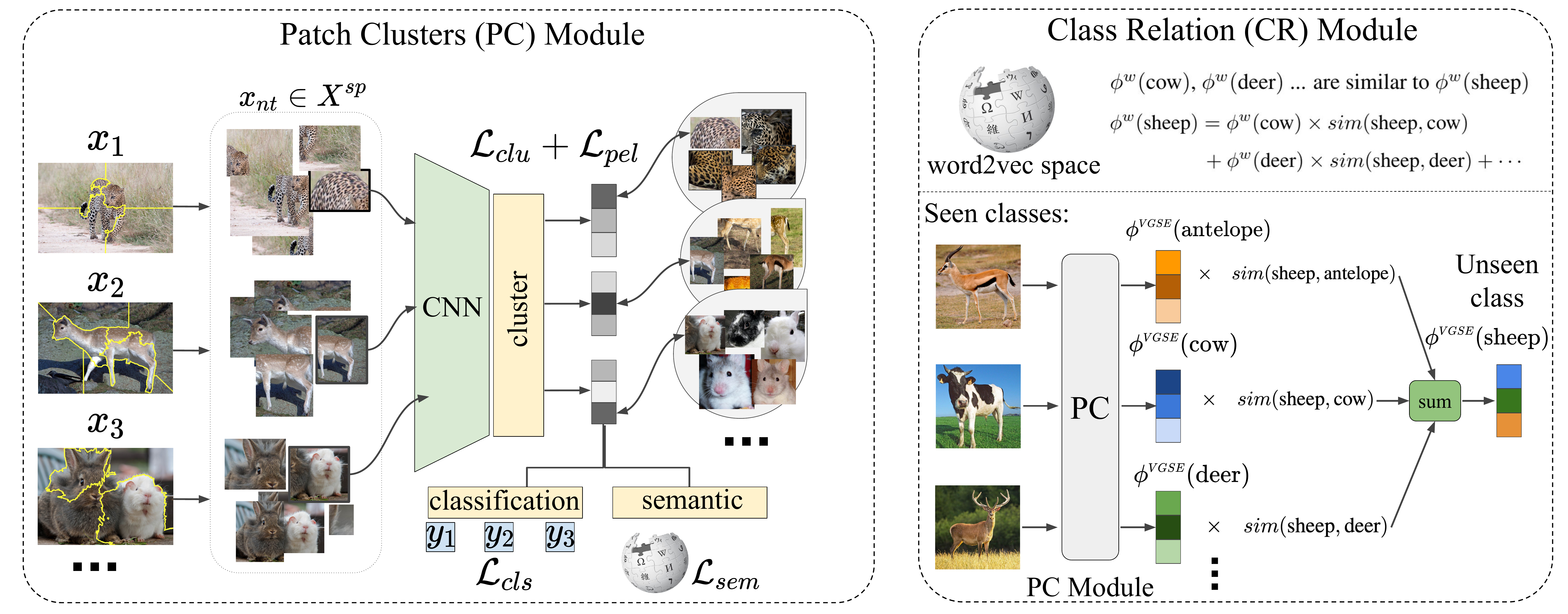}
    \caption{
        Our visually-grounded semantic embedding network consists of two modules. The Patch Clustering (PC) module learns clusters from patch images, and predicts semantic embeddings for seen classes with their images. The Class Relation (CR) module predicts the unseen class embeddings $\phi^{\Scale[0.5]{VGSE}}(y_m)$ using unseen and seen class relations learned from external knowledge, e.g., word2vec. For instance, the embedding for unseen class \textit{sheep} is predicted using the semantic embeddings of the seen classes, e.g., \textit{antelope}, \textit{cow}, \textit{deer}, and so on.
    }
  \label{fig:VGSE}
\end{figure*}

\myparagraph{Semantic Embeddings with Minimal Supervision} is drawing attention in image classification~\cite{kampffmeyer2019rethinking,bucher2017generating,rastegari2012attribute,bergamo2011picodes,sicre2017unsupervised}, transfer learning~\cite{chen2015deep,vittayakorn2016automatic,peng2017joint} and low-shot learning problems~\cite{al2017automatic,song2018selective,jiang2017learning,sharmanska2012augmented,nigam2019towards,yu2013designing}. Semantic embeddings collected from text corpora are alternatives to manual annotations, which include word embeddings learned from large corpora~\cite{yamada2018wikipedia2vec,glove,socher2013zero,w2v}, semantic relations such as knowledge graphs~\cite{wang2018zero,kampffmeyer2019rethinking,bucher2017generating}, and semantic similarities~\cite{cilibrasi2007google,wu2011flickr}, etc.
More recently, \cite{qiao2016less,zhu2018generative,al2017automatic,rohrbach2010helps} collect attribute-class associations from online encyclopedia articles that describe each category. 
The semantic similarity can be encoded by a taxonomical hierarchy or by incorporating co-occurrence statistics of words within the document.
However, this may not reflect visual similarity, e.g., \textit{sheep} is semantically close to \textit{dog} since they often co-occur in online articles, while visually \textit{sheep} is closer to a \textit{deer}. 
We focus on discovering visually-grounded semantic embeddings in the image space, and further incorporate the semantic relations between classes into our semantic embedding for better zero-shot knowledge transfer. 

\myparagraph{Learning Visual Properties from Image Patches.} 
Previous attempts for discovering middle-level representations for classification include exploring image-level embeddings by learning binary codes or classeme representations~\cite{torresani2010efficient,rastegari2012attribute,bergamo2011picodes}, and further introducing humans in the loop to discover localized and nameable attributes~\cite{parikh2011interactively,duan2012discovering}. However, those methods discover properties depicted in the whole image, which might result in a combination of several semantics covering several objects~(parts) that are hard to interpret~\cite{parikh2011interactively}. 
Visual transformer~\cite{dosovitskiy2020image} and BagNets~\cite{brendel2019approximating} showed that image patches can work as powerful visual words conveying visual cues for class discrimination. 
Bag of visual words~(BOVW) models~\cite{csurka2004visual,sivic2003video} propose to cluster image patches to learn a codebook and form image representations. However, BOVW extracts hand-crafted features followed by k-means clustering, while we learn clustering in an end-to-end manner via deep neural networks.
Considering the above problem, we propose to learn visual properties by clustering image patches, and predict the semantic embeddings with the visual properties depicted by patch clusters.

More closely related to our work are the ones learning discriminative image regions that can represent each class through clustering of local patches~\cite{singh2012unsupervised,doersch2013mid,doersch2012makes,sicre2017unsupervised}, e.g., finding representative elements to discriminate one class from others. Instead of picking up the most salient patches in each class, we aim to learn visual properties that are shared among different classes for most of the image patches appearing in the dataset. Besides, unlike some above methods that divide an image into a grid of square patches, we propose to use segmentation-based region proposals to obtain semantic image regions~(e.g., the entire head could represent one semantic region).


\section{Visually-Grounded Semantic Embedding}

We are interested in the (generalized) zero-shot learning task where the training and test classes are disjoint sets. 
The training set $\{\left ( x_n, y_n \right )  | x_{n} \in X^{s}, y_{n} \in Y^{s}\} _{n=1}^{N_s}$ consists of images $x_n$ and their labels $y_{n}$ from the seen classes $Y^{s}$. 
In the ZSL setting, test images are classified into unseen classes $Y^{u}$, and in the GZSL setting, into both $Y^{s}$ and $Y^{u}$ with the help of a semantic embedding space, e.g., human annotated attributes.
Since human-annotated attributes are costly to obtain, while prior unsupervised semantic embeddings are incomplete to describe the rich visual world, we propose to automatically discover a set of $D_{v}$ visual clusters as the semantic embedding, denoted by $\Phi^{\Scale[0.5]{VGSE}}  \in \mathbb{R}^{(|Y^{u}|+|Y^{s}|) \times D_{v}}$. 
The semantic embeddings for seen classes $\left \{\phi^{\Scale[0.5]{VGSE}} \left ( y \right ) | y\in Y^{s}  \right \} $, describing diverse visual properties of each category, are learned on seen classes images $X^{s}$. The semantic embeddings for unseen classes $\left \{\phi^{\Scale[0.5]{VGSE}} \left ( y \right ) | y\in Y^{u}  \right \}$ is predicted with the help of unsupervised word embeddings, e.g., w2v embeddings for class names $\Phi^{w} \in \mathbb{R}^{(|Y^{u}|+|Y^{s}|) \times D_{w}}$. 

Our Visually-Grounded Semantic Embedding~(VGSE) Network (see Figure~\ref{fig:VGSE}) consists of two main modules. (1) The Patch Clustering~(PC) module takes the training dataset as input, and clusters the image patches into $D_v$ visual clusters. Given one input image $x_n$, PC can predict the cluster probability $a_n \in \mathbb{R}^{D_v}$ indicating how likely the image would contain the visual property appearing in each cluster. (2) Since unseen class images cannot be observed during training, we propose the Class Relation~(CR) module to infer the semantic embeddings of unseen classes.
Finally, the learned semantic embedding $\Phi^{\Scale[0.5]{VGSE}}$ can be used to perform downstream tasks, e.g., Zero-Shot Learning.

\subsection{Patch Clustering (PC) Module}

\myparagraph{Patch image generation.} 
Patch-level embeddings allow us to explore the visual properties that appear in local image regions~\cite{26_wah2011caltech,dosovitskiy2020image}, e.g., the shape and texture of animal body parts or the objects in scenes. To obtain image patches that cover the entire semantic image region~(e.g. an animal head), we segment an image into regularly shaped regions via an unsupervised compact watershed segmentation algorithm~\cite{neubert2014compact}.
As shown in Figure~\ref{fig:VGSE}, for each image $x_n$, we find the smallest bounding box that fully covers each segment and crop $x$ into $N_t$ patches $\{ x_{nt} \}_{t=1}^{N_t}$ that cover different parts of the image. 
The number of patches $N_t$ is empirically set to be around $9$, as we observed in initial experiments that larger patches may include too many attributes, while smaller patches will be too tiny to contain any visual attribute.
In this way, we reconstruct our training set consisting of image patches $\{\left ( x_{nt}, y_n \right )  | x_{nt} \in X^{sp}, y_{n} \in Y^{s}\} _{n=1}^{N_s}$, here $\left | X^{sp} \right | =N_s N_t$, and $N_s$ is the train set size.

\myparagraph{Patch clustering.}
Our patch clustering module is a differentiable middle layer,
that simultaneously learns image patch representations and clustering. 
As shown in Figure~\ref{fig:VGSE} (left), we start from a deep neural network that extracts patch feature  $\theta \left ( x_{nt} \right ) \in \mathbb{R}^{D_f}$, where we use a ResNet~\cite{he2016deep} pretrained on ImageNet~\cite{39_Imagenet} as in other ZSL models~\cite{xian2018zero,xian2019}. Afterwards, a clustering layer $H: \mathbb{R}^{D_f} \to \mathbb{R}^{D_v} $ converts the feature representation into cluster scores:
\begin{equation}
       a_{nt} =  H \circ \theta \left ( x_{nt} \right )  \,,
\end{equation}
where $a_{nt}^k$ (the $k$-th element of $a_{nt}$) indicates the probability of assigning image patch $x_{nt}$ to cluster $k$, e.g., the patch clusters of spotty fur, fluffy head in Figure~\ref{fig:VGSE}.

A pretext task can be adopted to obtain semantically meaningful representations~\cite{van2020scan,noroozi2016unsupervised,he2020momentum} in an unsupervised manner. Our pretext task~\cite{van2020scan} enforces the image patch $x_{nt}$ and its neighbors being predicted to the same clusters. We retrieve nearest patch neighbors of $x_{nt}$ as $X_{nb}^{sp}$ by the $\mathcal{L}_2$ distance of patch features $\left \| \theta\left ( {x_{nt}} \right )  - \theta\left ( {x_i} \right )  \right \|_2 $, where $x_i\in X^{sp}$ and $x_i \neq x_{nt}$. 
The clustering loss is defined as
\begin{equation}
    \mathcal{L}_{clu} = - \sum_{x_{nt}\in X^{sp}} \sum_{x_{i} \in X_{nb}^{sp}} log(a_{nt}^T a_{i}) \,,
\end{equation}
where $a_{i} = H \circ \theta \left ( x_{i} \right ) $.
$\mathcal{L}_{clu}$ imposes consistent cluster assignment for $x_{nt}$ and its neighbors. To avoid all images being assigned to the same cluster, we follow\cite{van2020scan} to add an entropy penalty as follows:
\begin{equation}
   \mathcal{L}_{pel} = \sum_{k=1}^{D_v} \bar{a}_{nt}^k \log \bar{a}_{nt}^k  \,, \quad 
   \bar{a}_{nt}^k = \frac{1}{N_s N_t }  \sum_{x_{nt} \in X^{sp}} {a_{nt}^k}  \,,
\end{equation}
ensuring that images are spread uniformly over all clusters.

\myparagraph{Class discrimination.} To impose class discrimination information into the learnt clusters, we propose to apply an cluster-to-class layer $Q: \mathbb{R}^{D_v} \to \mathbb{R}^{|Y^s|}$ to map the cluster prediction of each image to the class probability, i.e., $p(y|x_{nt}) = softmax \left (Q \circ \theta \left ( x_{nt} \right ) \right ) $. We train this module with the following cross-entropy loss,
\begin{equation}
    \mathcal{L}_{cls} = -\log\frac{\exp \left ( p\left ( y_{n} | x_{nt} \right ) \right ) }{\sum_{\hat{y} \in Y^s}{\exp \left ( p\left ( \hat{y} | x_{nt} \right ) \right )} } \,.
\end{equation}

\myparagraph{Semantic relatedness.} 
We further encourage the learned visual clusters to be transferable between classes, to benefit the downstream zero-shot learning tasks. We learn clusters shared between semantically related classes, e.g., \textit{horse} share more semantic information with \textit{deer} than with \textit{dolphin}. We implement this by mapping the learned cluster probability to the semantic space constructed by w2v embeddings $\Phi^w$. The cluster-to-semantic layer $S: \mathbb{R}^{D_v} \to \mathbb{R}^{D_w}$ is trained by regressing the w2v embedding for each class,
\begin{equation}
   \mathcal{L}_{sem} =  \left \| S \circ a_{nt} - \phi^{w}(y_n) \right \|_2 \,, 
\end{equation}
where $y_n$ denotes the ground truth class, and $\phi^w \left (y_{n} \right ) \in \mathbb{R}^{D_w}$ represents the w2v embedding for the class $y_n$.

The overall objective for training the model is as follows:
\begin{equation}
    \mathcal{L} = \mathcal{L}_{clu} + \lambda \mathcal{L}_{pel} + \beta  \mathcal{L}_{cls} + \gamma  \mathcal{L}_{sem} \,.
\end{equation}

\myparagraph{Predict seen semantic embeddings.} After we learned the visual clusters, given one input image patch $x_{nt}$, the model extracts the feature $\theta\left ( x_{nt} \right )$ followed by predicting the cluster probability $a_{nt} = H \circ \theta\left ( x_{nt} \right ) \in \mathbb{R}^{D_v}$ where each dimension indicates the likelihood that the image patch $x_{nt}$ being assigned to a certain cluster learned by this module.

The image embedding $a_n \in \mathbb{R}^{D_v}$ for $x_{n}$ is calculated by averaging the patch embedding in that image:
\begin{equation}
\label{eq:7}
    a_n = \frac{1}{N_t} \sum_{t=1}^{N_t} a_{nt} \,.
\end{equation}

Similarly, we calculate the semantic embedding for $y_n$ by averaging the embeddings of all images belonging to $y_n$:
\begin{equation}
\label{eq:seen_att}
    \phi^{\Scale[0.5]{VGSE}}(y_n) = \frac{1}{|I_i|}\sum_{j \in I_i} a_j \,,
\end{equation}
%
%
where 
$I_i$ is the indexes of all images belonging to class $y_n$,
and $a_j$ denotes the image embedding of the $j$-th image.

\subsection{Class Relation (CR) Module}

While seen semantic embeddings can be estimated from training images using Eq.~\ref{eq:seen_att}, how to compute the unseen semantic embeddings is not straightforward since their training images are not available. 
As semantically related categories share common properties, e.g., \textit{sheep} and \textit{cow} both live on grasslands, 
we propose to learn a Class Relation Module to formulate the similarity between seen classes $Y^{s}$ and unseen classes $Y^{u}$.
In general, any external knowledge, e.g., word2vec~\cite{w2v,glove} or human-annotated attributes, can be utilized to formulate the relationship between two classes. Here we use word2vec learned from a large online corpus to minimize the human annotation effort.
Below, we present two solutions to learn the class relations: (1) directly averaging the semantic embeddings from the neighbor seen classes in the word2vec spaces, (2) optimizing a similarity matrix between unseen and seen classes.

\myparagraph{Weighted Average~(WAvg).}  For unseen class $y_m$, we first retrieve several nearest class neighbours in seen classes  
by the similarity measured with $\mathcal{L}_2$ distance over w2v embedding space
, and we denote the neighbor classes set as $Y^{s}_{nb}$. The semantic embedding vector for $y_m$ is calculated as the weighted combination~\cite{al2016recovering} of seen semantic embeddings:
\begin{equation}
\begin{aligned}
   \phi^{\Scale[0.5]{VGSE}}(y_m) = \frac{1}{|Y^{s}_{nb}| } \sum_{\tilde{y} \in Y^{s}_{nb}}   sim\left ( y_m, \tilde{y} \right )  \cdot \phi^{\Scale[0.5]{VGSE}}(\tilde{y}) \,, 
\end{aligned}
\label{equ:wavg}
\end{equation}
\begin{equation}
    sim(y_m, \tilde{y}) = \exp( - \eta   \left \| \phi^{w}\left ( y_m \right )  - \phi^{w}\left ( \tilde{y} \right ) \right \|  _2 ) \,,
\end{equation}
where $\text{exp}$ stands for the exponential function and $\eta$ is a hyperparameter to adjust the similarity weight. We denote our semantic embeddings learned with weighted average strategy as \texttt{VGSE-WAvg}.

\myparagraph{Similarity Matrix Optimization~(SMO).} Given the w2v embeddings $\phi^{w}({Y}^{s}) \in \mathbb{R}^{|Y^{s}| \times D_w}$ of seen classes and embedding $\phi^{w}(y_m)$ for unseen class $y_m$, we learn a similarity mapping $r \in \mathbb{R}^{|Y^{s}|}$, where $r_{i}$ denotes the similarity between the unseen class $y_m$ and the $i$-th seen class.
 The similarity mapping is learned via the following optimization problem:
\begin{equation}
\begin{aligned}
   & \min _{r} \left \| \phi^{w}(y_m) - r^T \phi^{w}(Y^{s}) \right \| _2\\
   & \quad \textrm{s.t.} \quad \alpha < r < 1 \quad and \quad \sum_{i=1}^{|Y^s|}r_{i}  = 1 .
\end{aligned}
\label{equ:class-relation}
\end{equation}
Here $\alpha$ is the lower bound which can be either $0$ or $-1$, indicating whether we only learn positive class relations or we learn negative relations as well. 
We base this mapping on the assumption that semantic embeddings follow linear analogy, e.g., $\phi^{w}(\text{king}) - \phi^{w}(\text{man}) + \phi^{w}(\text{woman}) \approx  \phi^{w}(\text{queen})$, which holds for w2v embeddings and our semantic embeddings $\phi^{\Scale[0.5]{VGSE}}$. 
After the mapping is learned, we can predict the semantic embeddings for the unseen class $y_m$ as:
\begin{equation}
    \phi^{\Scale[0.5]{VGSE}}(y_m) = r^T \phi^{\Scale[0.5]{VGSE}}(Y_s) \,,
\end{equation}
where the value of each discovered semantic embedding for unseen class $y_m$ is the weighted sum of all seen class semantic embeddings. We denote our semantic embeddings learned with similarity matrix optimization~(SMO) as \texttt{VGSE-SMO}.

\setlength{\tabcolsep}{4pt}
\renewcommand{\arraystretch}{1.2} 
\begin{table*}[t]
\centering
\small
 \resizebox{.85\linewidth}{!}{%
   \begin{tabular}{l | l | l | c c c |c c c |c c c |c c c }
  	& & & \multicolumn{3}{c|}{\textbf{Zero-Shot Learning}} & \multicolumn{9}{c}{\textbf{Generalized Zero-Shot Learning}} \\
  	& & & \textbf{AWA2} & \textbf{CUB} & \textbf{SUN} & \multicolumn{3}{c}{\textbf{AWA2}} & \multicolumn{3}{c}{\textbf{CUB}} & \multicolumn{3}{c}{\textbf{SUN}}  \\
  & \textbf{ZSL Model}	& \textbf{Semantic Embeddings} & \textbf{T1} & \textbf{T1} & \textbf{T1}& \textbf{u} & \textbf{s} & \textbf{H} & \textbf{u} & \textbf{s} & \textbf{H} & \textbf{u} & \textbf{s} & \textbf{H}  \\
  	
  	\hline
  	
  	\multirow{4}{2em}{\rotatebox{90}{Generative}} 
  	& \multirow{2}{8em}{\texttt{CADA-VAE}~\cite{schonfeld2019}}
  	& \texttt{w2v}~\cite{w2v} & 49.0 & 22.5 & 37.8 & 38.6 & 60.1 & 47.0 & 16.3 & 39.7 & 23.1 & 26.0 & 28.2 & 27.0 \\
  	& & \texttt{VGSE-SMO}~(Ours) & 52.7 & 24.8 & 40.3 & 46.9 & 61.6 & 53.9 & 18.3 & 44.5 & 25.9 & \textbf{29.4} & 29.6 & 29.5 \\
  		\cline{2-15}
  	&\multirow{2}{8em}{\texttt{f-VAEGAN-D2}~\cite{xian2019}}
  	& \texttt{w2v}~\cite{w2v} & 58.4 & 32.7 & 39.6 & 46.7 & 59.0 & 52.2 & 23.0 & 44.5 & 30.3 & 25.9 & 33.3 & 29.1 \\
  	& & \texttt{VGSE-SMO}~(Ours) & 61.3 & \textbf{35.0} & \textbf{41.1} & 45.7 & 66.7 & 54.2 & \textbf{24.1} & 45.7 & \textbf{31.5} & 25.5 & \textbf{35.7} & \textbf{29.8}\\
  		\hline
  	
  	\multirow{6}{2em}{\rotatebox{90}{Non-Generative }} & \multirow{2}{8em}{\texttt{SJE}~\cite{akata2015evaluation}}
  	& \texttt{w2v}~\cite{w2v} & 53.7 & 14.4 & 26.3 & 39.7 & 65.3 & 48.8 & 13.2 & 28.6 & 18.0 & 19.8 & 18.6 & 19.2 \\
  	& & \texttt{VGSE-SMO}~(Ours) & 62.4 & 26.1 & 35.8 & 46.8 & 72.3 & 56.8 & 16.4 & 44.7 & 28.3 & 28.7 & 25.2 & 26.8 \\
  	\cline{2-15}
  	
    & \multirow{2}{8em}{\texttt{GEM-ZSL}~\cite{liu2021goal}}
  	& \texttt{w2v}~\cite{w2v} & 50.2 & 25.7 & - & 40.1 & 80.0 & 53.4 & 11.2 & \textbf{48.8} & 18.2 & - & - & - \\
  	
  	& & \texttt{VGSE-SMO}~(Ours) & 58.0 & 29.1 & - & 49.1 & 78.2 & 60.3 & 13.1 & 43.0 & 20.0 & - & - & -\\
  		\cline{2-15}
  	& \multirow{2}{8em}{\texttt{APN}~\cite{xu2020attribute}}
  	& \texttt{w2v}~\cite{w2v} & 59.6 & 22.7 & 23.6 & 41.8 & 75.0 & 53.7 & 17.6 & 29.4 & 22.1 & 16.3 & 15.3 & 15.8 \\
  	& & \texttt{VGSE-SMO}~(Ours) & \textbf{64.0} & 28.9 & 38.1 & \textbf{51.2} & \textbf{81.8} & \textbf{63.0}  & 21.9 & 45.5 & 29.5 & 24.1 & 31.8 & 27.4 \\
  \end{tabular}
}
\caption{Comparing our \texttt{VGSE-SMO}, with \texttt{w2v} semantic embedding over state-of-the-art ZSL models. In ZSL, we measure Top-1 accuracy~(\textbf{T1}) on unseen classes, in GZSL on seen/unseen~(\textbf{s/u}) classes and their harmonic mean~(\textbf{H}). Feature Generating Methods, i.e., \texttt{f-VAEGAN-D2}, and \texttt{CADA-VAE} generating synthetic training samples, and \texttt{SJE}, \texttt{APN}, \texttt{GEM-ZSL} using only real image features. 
}
\label{tab:ZSL_acc}
\end{table*}
\section{Experiments}

After introducing the datasets and experimental settings, we demonstrate that our \texttt{VGSE} outperforms unsupervised word embeddings over three benchmark datasets and this phenomenon generalizes to five SOTA ZSL models~(\S\ref{sec:sota}). With extensive ablation studies, we showcase clustering with images patches is effective for learning the semantic embeddings, and demonstrate the effectiveness of the PC module and CR module~(\S\ref{sec:ablation}). In the end, we present visual clusters as qualitative results~(\S\ref{sec:quali}, \S\ref{sec:human}).

\myparagraph{Dataset.}
We validate our model on three ZSL benchmark datasets. AWA2~\cite{xian2018zero} is a coarse-grained dataset for animal categorization, containing $30,475$ images from $50$ classes, where $40$ classes are seen and $10$ are unseen classes.
CUB~\cite{26_wah2011caltech} is a fine-grained dataset for bird classification, containing $11,788$ images and  $200$ classes, where $150$ classes are seen and $50$ are unseen classes.
SUN~\cite{25_SUNdataset} is also a fine-grained dataset for scene classification, with $14,340$ images coming from $717$ scene classes, where $645$ classes are seen and $72$ are unseen classes.

\myparagraph{Implementation details.} Specifically, in the patch clustering (PC) module we learn seen-semantic embeddings with train set~(seen classes) proposed by ~\cite{xian2018zero}, the unseen-class embeddings are predicted in the class relation (CR) module without seeing unseen images. We adopt ResNet50~\cite{he2016deep} pretrained on ImageNet1K~\cite{39_Imagenet} as the backbone.
The cluster number $D_v$ is set as $150$ for three datasets. 
For the Weighted Average module in Eq.~\ref{equ:wavg}, we set $\eta$ as 5 for all datasets, and use $5$ neighbors for all datasets. For the similarity matrix optimization in Eq.~\ref{equ:class-relation}, we set $\alpha$ as -1 for AWA2 and CUB, and as 0 for SUN. 
More details are in the supplementary.

\myparagraph{Semantic embeddings for ZSL.}
To be fair, we compare our \texttt{VGSE} semantic embeddings with other alternatives using the same image features and ZSL models. All the image features are extracted from ResNet101~\cite{he2016deep} pretrained on ImageNet~\cite{39_Imagenet}. We follow the data split provided by~\cite{xian2018zero}. The semantic embeddings are L2 normalized following~\cite{xian2018zero}. All ablation studies use the \texttt{SJE}~\cite{akata2015evaluation,xu2020attribute} as the ZSL model as it is simple to train.
Besides, we verify the generalization ability of our semantic embeddings over five state-of-the-art ZSL models with their official code.
The non-generative models include \texttt{SJE}~\cite{akata2015evaluation}, \texttt{APN}~\cite{xu2020attribute}, \texttt{GEM-ZSL}~\cite{liu2021goal}, learning a compatibility function between image and semantic embeddings. The generative approaches consist of \texttt{CADA-VAE}~\cite{schonfeld2019} and \texttt{f-VAEGAN-D2}~\cite{xian2019}, learning a generative model that synthesizes image features of unseen classes from their semantic embeddings.
Note that for all ZSL models, we use the same hyperparameters as proposed in their original papers for all semantic embeddings with no hyperparameter tuning.

\subsection{Comparing with the State-of-the-Art}
\label{sec:sota}
We first compare our semantic embeddings \texttt{VGSE-SMO} with the unsupervised word embeddings \texttt{w2v}~\cite{w2v} on three benchmark datasets and five ZSL models. 
We further compare ours with other state-of-the-art methods that learn semantic embeddings with less human annotation.

\myparagraph{\texttt{VGSE} surpasses \texttt{w2v} by a large margin.}
The results shown in Table~\ref{tab:ZSL_acc} demonstrate that our \texttt{VGSE-SMO} semantic embeddings significantly outperform word embedding \texttt{w2v} on all datasets and all ZSL models.  
Considering the non-generative ZSL models, \texttt{VGSE-SMO} outperform \texttt{w2v} on all three datasets by a large margin. In particular, on AWA2 dataset, when coupled with \texttt{GEM-ZSL}, our \texttt{VGSE-SMO} boosts the ZSL performance of \texttt{w2v} from 50.2\% to 58.0\%. On the fine-grained datasets CUB and SUN, \texttt{VGSE-SMO} achieves even higher accuracy boosts. For example, when coupled with the \texttt{APN} model,  \texttt{VGSE-SMO} increases the ZSL accuracy of CUB from 22.7\% to 28.9\%, and the accuracy of SUN from 23.6\% to 38.1\%. These results demonstrate that our approach not only works well on generic object categories, but also has great potential to benefit the challenging fine-grained classification task. \texttt{VGSE} improves the GZSL performance of both seen and unseen classes, yielding a much better harmonic mean~(e.g., when trained with \texttt{SJE}, \texttt{VGSE-SMO} improves over the harmonic mean of \texttt{w2v} by 8.0\% on AWA2, 10.3\% on CUB, and 7.6\% on SUN). These results indicate that our \texttt{VGSE} facilitates the model to learn a better compatibility function between image and semantic embeddings, for both seen and unseen classes.

Our \texttt{VGSE} semantic embeddings show great potential on generative models as well. In particular, \texttt{VGSE} coupled with f-VAEGAN-D2 surpasses all other methods by a wide margin on SUN and CUB datasets, i.e., we obtain 35.0\% vs 32.7\%~(\texttt{w2v}) on CUB, and 41.1\% vs 39.6\%~(\texttt{w2v}) on SUN. 
As our embeddings are more machine detectable than \texttt{w2v}, introducing visual properties to the conditional GAN will allow them to generate more discriminative image features. 

\begin{table}[tb]
\centering
\resizebox{\linewidth}{!}{
\begin{tabular}{l | c |c c c}
\toprule
\multirow{2}{*}{\textbf{Semantic Embeddings}} & \textbf{External} & \multicolumn{3}{c}{\textbf{Zero-shot learning}} \\
 & \textbf{knowledge} & \textbf{AWA2}        & \textbf{CUB}         & \textbf{SUN}        \\ \hline
\texttt{w2v}~\cite{w2v} & w2v & 58.4 & 32.7 & 39.6 \\
\texttt{ZSLNS}~\cite{qiao2016less} &   $T$                              & 57.4        & 27.8        & -          \\
\texttt{GAZSL}~\cite{zhu2018generative} & $T$                                     & -           & 34.4        & -          \\
\texttt{Auto-dis}~\cite{al2017automatic} &  $T$                                    &  52.0       & -        & -       \\
\texttt{CAAP}~\cite{al2016recovering} &  $T$ and $H$                                    & 55.3        & 31.9        & 35.5       \\ \hline
\texttt{VGSE-SMO}~(Ours)                          & w2v                                 & \textbf{61.3} $\pm$ 0.3        & \textbf{35.0 }  $\pm$ 0.2      &\textbf{41.1}  $\pm$ 0.3      \\
\bottomrule
\end{tabular}}
\caption{Comparing with state-of-the-art methods for learning semantic embeddings with less human annotation (T: online textual articles, H: human annotation) using same image features and ZSL model~(\texttt{f-VAEGAN-d2}~\cite{xian2019}). }
\label{tab:SOTA}
\end{table}

\myparagraph{\texttt{VGSE} outperforms SOTA weakly supervised ZSL semantic embeddings.}  
We compare \texttt{VGSE} with other works that learn ZSL semantic embeddings with less human annotation. \texttt{CAAP}~\cite{al2016recovering} learns the unseen semantic embeddings with the help of w2v and the human annotated attributes for seen classes. \texttt{Auto-Dis}~\cite{al2017automatic} collects attributes from online encyclopedia articles that describe each category, and learn attribute-class association with the supervision of visual data and category label. \texttt{GAZSL}~\cite{zhu2018generative} and \texttt{ZSLNS}~\cite{qiao2016less} learn semantic embeddings from wikipedia articles.

The results shown in Table~\ref{tab:SOTA} demonstrate that our \texttt{VGSE} embedding, using only w2v as external knowledge, surpasses all other method that uses textual articles on three datasets.  In particular, our \texttt{VGSE-SMO} achieves an accuracy of 61.3\% on AWA2, improving the closest semantic embedding \texttt{w2v} by 2.9\%. On SUN, we also outperform the closest semantic embedding \texttt{w2v} by 1.5\%.

\subsection{Ablation study}
\label{sec:ablation}

We provide ablation studies for our PC and CR modules. 

\begin{table}[tb]
\centering
\small
\resizebox{\linewidth}{!}{
\begin{tabular}{l |c c c}
\toprule
 \multirow{2}{*}{\textbf{Semantic Embeddings}} & \multicolumn{3}{c}{\textbf{Zero-shot learning}} \\
 & AWA2        & CUB         & SUN        \\ \hline
\texttt{k-means-SMO} & 54.5 $\pm$ 0.4 & 15.0 $\pm$ 0.5 & 25.2 $\pm$ 0.4  \\ 
\texttt{ResNet-SMO} & 55.3 $\pm$ 0.2 & 15.4 $\pm$ 0.1 & 25.1 $\pm$ 0.1\\
\hline
 $\mathcal{L}_{clu} + \mathcal{L}_{pel}$~(baseline + \texttt{SMO}) & 56.6 $\pm$ 0.2 & 16.7 $\pm$ 0.2 &  26.3 $\pm$ 0.3    \\
$\qquad \quad + \mathcal{L}_{cls}$ & 61.2 $\pm$ 0.1 & 23.7 $\pm$ 0.2 &  30.5 $\pm$ 0.2 \\ 
$\qquad \qquad + \mathcal{L}_{sem}$~(\texttt{VGSE-SMO}) & \textbf{62.4} $\pm$ 0.3 & \textbf{26.1} $\pm$ 0.3 & \textbf{35.8} $\pm$ 0.2   \\ \hline
\texttt{VGSE-WAvg} & 57.7 $\pm$ 0.2 & 25.8 $\pm$ 0.3 & 35.3 $\pm$ 0.2   \\

\bottomrule
\end{tabular}}
\caption{Ablation study over the \texttt{PC} module reporting ZSL T1 on AWA2, CUB, and SUN~(mean accuracy and std over 5 runs). The baseline is the \texttt{PC} module with the cluster loss $\mathcal{L}_{clu}$ and $\mathcal{L}_{pel}$.
Our full model \texttt{VGSE-SMO} is trained with two additional losses $\mathcal{L}_{cls}$, $\mathcal{L}_{sem}$. Two kinds of semantic embeddings learned from k-means clustering and pretrained ResNet are listed below for comparison.}
\label{tab:clustering_ablation}
\end{table}

\myparagraph{Is PC module effective?}
We first ask if learning semantic embeddings through clustering is effective in terms of ZSL accuracy, when compared to other alternatives.
We compare our semantic embeddings against the following baselines:

\textit{ResNet features} are extracted by feeding image patch $x_{nt}$ to a pretrained ResNet50. 
We follow Eq.~\ref{eq:7} and Eq.~\ref{eq:seen_att} to predict semantic embeddings for seen classes. 
\textit{K-means clustering} is an alternative for our clustering model. We cluster the patch images features $\theta{(x_{nt})}$ learned from our \texttt{PC} module into $D_v$ visual clusters. The patch embedding $a_{nt}^k$ is defined as the cosine similarity between the patch feature $\theta({x_{nt}})$ and the cluster center. In both cases the unseen semantic embeddings are predicted with our \texttt{SMO} module.

We ablate our losses and compare our \texttt{VGSE-SMO} with the two alternatives,  
then report ZSL results on three benchmark datasets in Table~\ref{tab:clustering_ablation}.
First, the \texttt{k-means-SMO} achieves on par results with our baseline model trained with only the cluster losses $\mathcal{L}_{clu}$ and $\mathcal{L}_{pel}$~\cite{van2020scan}, the reason we adopt~\cite{van2020scan} instead of k-means is that we can easily train the network with our proposed losses in an end-to-end manner. Second, the addition of the classification loss $\mathcal{L}_{cls}$ leads to notable improvement over the baseline model trained with $\mathcal{L}_{clu}$ and $\mathcal{L}_{pel}$, and the semantic relatedness loss $\mathcal{L}_{sem}$ further improve the performance of our semantic embeddings, e.g., in total, we gain $5.8\%$, $9.4\%$ and $9.5\%$ improvement on AWA2, CUB, and SUN, respectively. The result demonstrates that imposing class discrimination and semantic relatedness leads to better performance in the ZSL setting. Third, our \texttt{VGSE-SMO} embeddings improve over the \texttt{ResNet-SMO} embeddings by $7.1\%$, $10.7\%$ and $10.7\%$ on AWA2, CUB, and SUN, respectively. We conjuncture that the visual clusters learned in our model is shared among different classes and lead to better generalization ability when the training and testing sets are disjoint~(see qualitative results in Figure~\ref{fig:teaser} and Section~\ref{sec:quali}).

\begin{figure}
\centering
 \resizebox{1\linewidth}{!}{
    \subfloat[cluster number $D_v$]{
    \includegraphics[width=.5\linewidth]{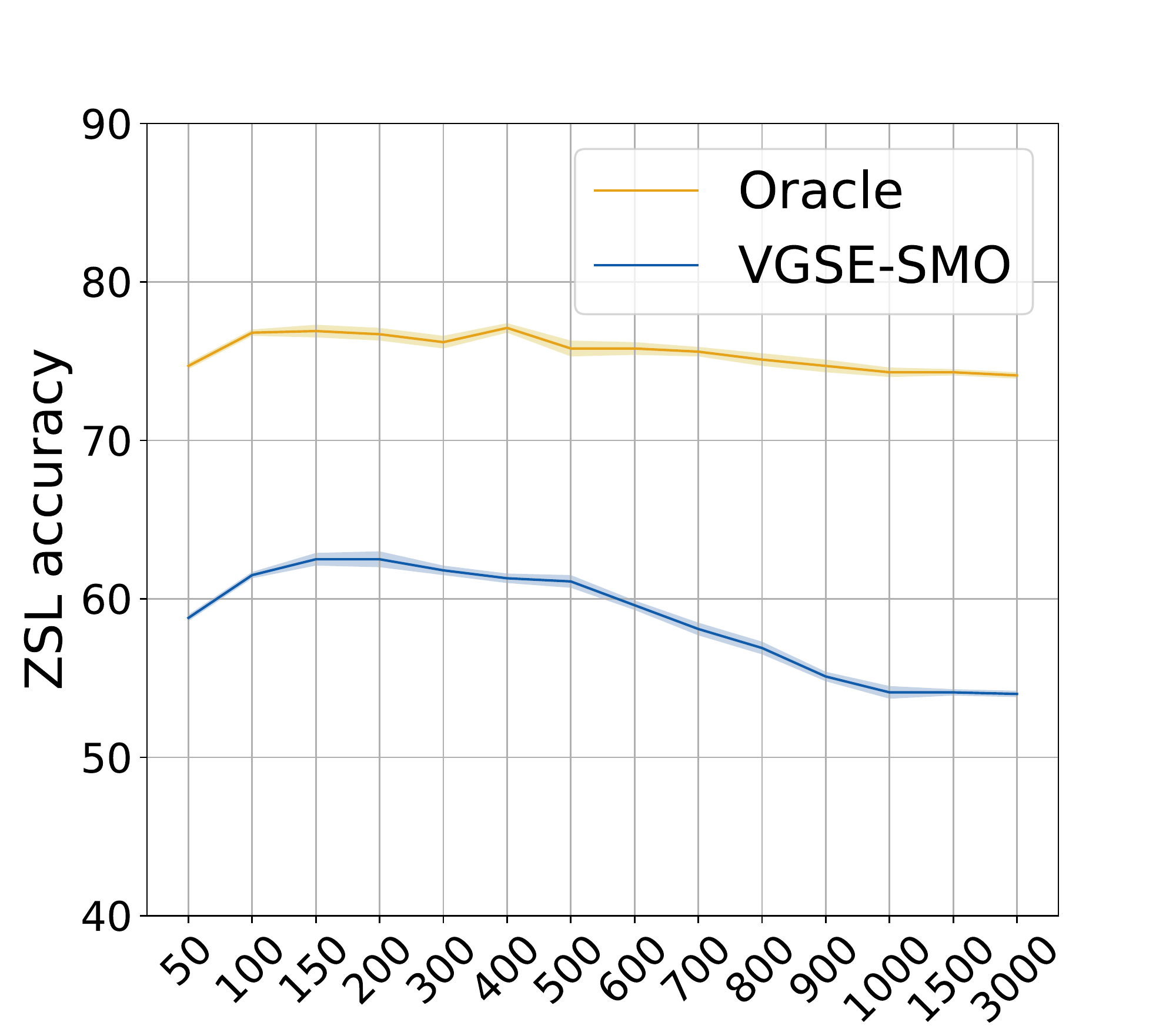}
    \label{fig:cluster_num}}
\hfill
    \subfloat[patch number $N_t$]{
    \includegraphics[width=.5\linewidth]{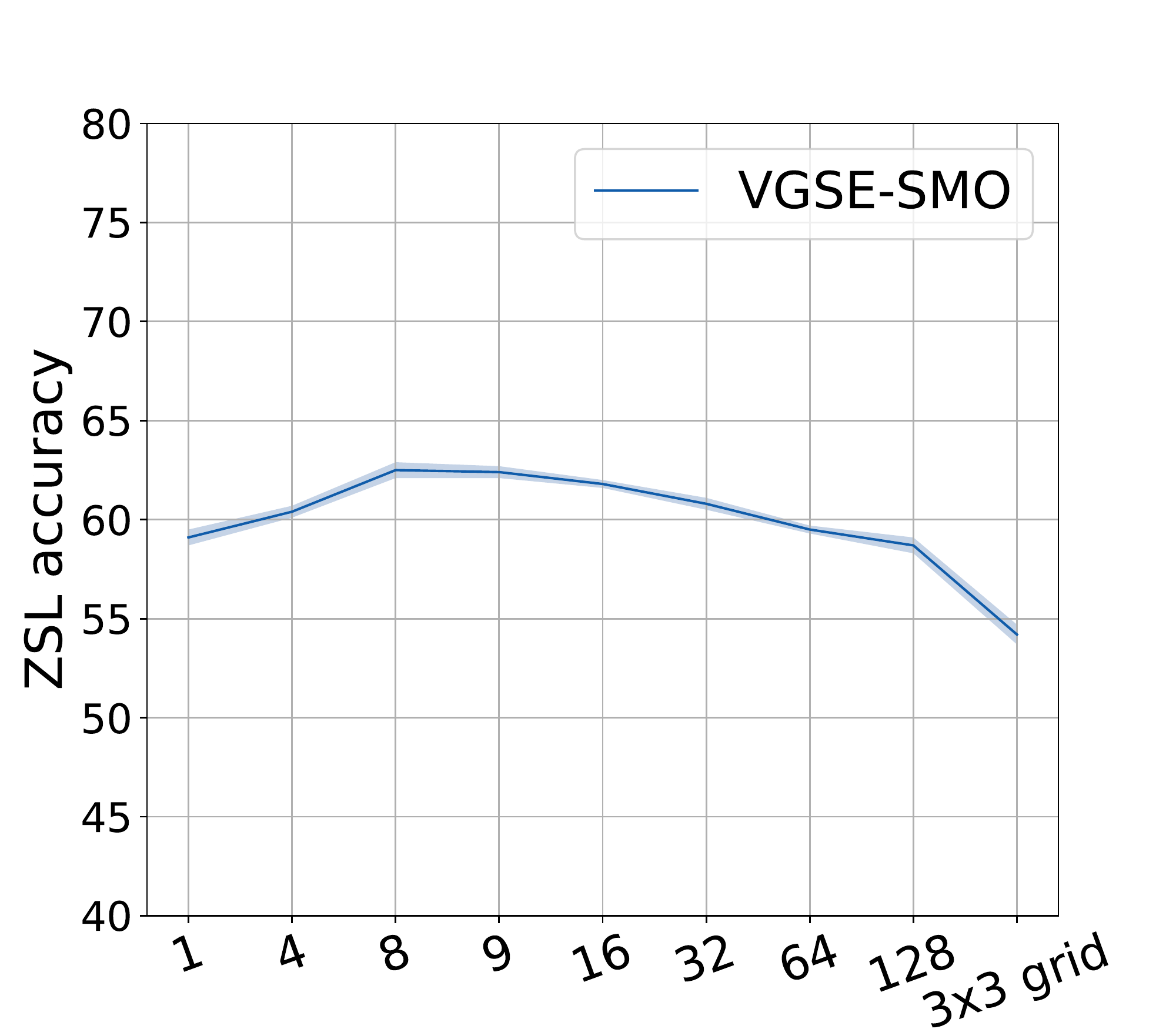}
    \label{fig:patch_num}}
    }
    \caption{(a) Influence of the cluster number $D_v = 50,\dots,3000$.
    In the oracle setting, we feed unseen classes images to the \texttt{PC} module to predict unseen semantic embeddings. (b) Influence of the patch number $N_t$ we used per image with the watershed segmentation for obtaining our \texttt{VGSE-SMO} class embeddings. $N_t = 1$ uses the whole image (no patches).  
    ``3$\times$3 grid'' 
    crops the image into 9 square patches. 
    Both plots report ZSL accuracy with 
    \texttt{SJE} model trained on AWA2 dataset (mean and std over 5 runs). 
    }
    
    \label{fig:cluster_patch}
\end{figure}

\myparagraph{How many clusters are needed?}
To measure the influence of the cluster number $D_v$ on our semantic embeddings, we train the \texttt{PC} module with various $D_v$~(results shown in Figure~\ref{fig:cluster_num}). When the unseen semantic embeddings are predicted under an oracle setting~(predicted from the unseen class images), various dimension $D_v$ does not influence the classification accuracy on unseen classes~(the orange curve). While under the ZSL setting where unseen semantic embeddings are predicted from class relations~(\texttt{VGSE-SMO}), the cluster numbers influence the ZSL performance.
Before the cluster number increases up to a breaking point~($D_v=200$), the ability of the semantic embeddings is also improved~(from $58.4\%$ to $62.5\%$), since the learned clusters contain visually similar patches from different classes, which can model the visual relation between classes. However, increasing the number of clusters leads to small pure clusters~(patches coming from one single category), resulting in poor generalization between seen and unseen classes.

\myparagraph{SMO vs WAvg.} We compare our two class relation functions \texttt{VGSE-WAvg} and \texttt{VGSE-SMO} in Table~\ref{tab:clustering_ablation}~(Row 7 and 6). The results demonstrate that \texttt{VGSE-WAvg} works on par with \texttt{VGSE-SMO} on SUN and CUB datasets, with $< 0.5\%$ performance gap. While on AWA2 dataset, \texttt{VGSE-SMO} yields better ZSL performance~(with $62.4\%$) than \texttt{VGSE-WAvg}~(with $57.7\%$). The results indicate that predicting the unseen semantic embeddings with the weighted average of a few seen classes semantic embeddings~(\texttt{VGSE-WAvg}) is working well for fine-grained datasets since the visual discrepancy between classes is small. However, for coarse-grained dataset AWA2, the class relation function considering all the seen classes embeddings~(\texttt{VGSE-SMO}) works better.

\myparagraph{Ablation over patches.}
We further study if using patches for clustering is better than using the whole image, and how many patches do we need from one image. The experiment results in Figure~\ref{fig:patch_num} demonstrate that with the patch number increase from $1$~(single image clustering) to $9$, the ZSL performance increases as well, since the image patches used for semantic embedding learning contain semantic object parts and thus result in better knowledge transfer between seen and unseen classes. However, for a large $N_t$, the patches might be 
too tiny to contain consistent semantic, thus resulting in performance dropping, e.g., the ZSL accuracy on AWA2 drops from 62.4\%~($N_t = 9$) to 58.7\%~($N_t = 128$). We also compare the patches generated by watershed segmentation proposal with using 3$\times$3 grid patches ($N_t = 9$),
and we found that using watershed as the region proposal results in accuracy boost~($8.2\%$ on AWA2) compared to the regular grid patch, since the former patches tend to cover more complete object parts rather than random cropped regions.

\begin{table}[tb]
\centering
\small
 \resizebox{.95\linewidth}{!}{
\begin{tabular}{l |c c | c c}
\toprule
 \multirow{2}{*}{\textbf{Semantic Embeddings}} & \multicolumn{2}{c}{\textbf{AWA2}} & \multicolumn{2}{c}{\textbf{CUB}} \\
& \textbf{T1} &   \textbf{H}      & \textbf{T1}        & \textbf{H}      \\ \hline
\texttt{w2v}~\cite{w2v} & 53.7 $\pm$ 0.2 & 48.8 $\pm$ 0.1 &  14.4 $\pm$ 0.3 & 18.0 $\pm$ 0.2   \\
\texttt{VGSE-SMO} (\texttt{w2v}) & 62.4 $\pm$ 0.1 & 56.8 $\pm$ 0.1 &  26.1 $\pm$ 0.2 & 28.3 $\pm$ 0.1  \\ \hline
\texttt{glove}~\cite{glove} & 38.8 $\pm$ 0.2  & 38.7 $\pm$ 0.3  & 19.3 $\pm$ 0.2 &  13.4 $\pm$ 0.1 \\
\texttt{VGSE-SMO} (\texttt{glove}) & 46.5 $\pm$ 0.1 & 46.0 $\pm$ 0.1 & 25.2 $\pm$ 0.3 &  27.1 $\pm$ 0.2 \\ \hline
\texttt{fasttext}~\cite{bojanowski2017enriching} & 47.7 $\pm$ 0.1 & 44.6 $\pm$ 0.3 & -  &  -  \\
\texttt{VGSE-SMO} (\texttt{fasttext}) & 51.9 $\pm$ 0.2 & 53.2 $\pm$ 0.1 & -  &  -  \\ \hline
\texttt{Attribute} &  62.8 $\pm$ 0.1 & 62.6 $\pm$ 0.3 &  56.4 $\pm$ 0.2 &  49.4 $\pm$ 0.1 \\
\texttt{VGSE-SMO} (\texttt{Attribute}) & 66.7 $\pm$ 0.1 &  64.9 $\pm$ 0.1 &  56.8 $\pm$ 0.1 &  50.9 $\pm$ 0.2 \\

\bottomrule
\end{tabular}}
\caption{Evaluating the external knowledge, i.e., word embeddings w2v~\cite{w2v}, glove~\cite{glove}, fasttext~\cite{bojanowski2017enriching}, and the human annotated attributes, for our \texttt{VGSE-SMO} embeddings, e.g., \texttt{VGSE-SMO} (\texttt{glove})  
indicates that \texttt{CR} module is trained with \texttt{glove} embedding. 
\textbf{T1}: top-1 accuracy in ZSL, \textbf{H}: harmonic mean in GZSL trained with \texttt{SJE}~\cite{akata2015evaluation} on AWA2, and CUB~(std over 5 runs). }
\label{tab:sideinfo_ablation}
\end{table}

\myparagraph{Can we do better with human annotated attributes?} 
Table~\ref{tab:sideinfo_ablation} shows the performance of our model when different external knowledge is used to perdict the unseen class embeddings in the \texttt{CR} module. Nearly all of our conclusions from former section carry over, e.g., \texttt{VGSE-SMO} class embeddings outperform the other class embeddings by a large margin. For instance, we improve the ZSL accuracy over \texttt{glove} by $7.7\%$~(AWA2) and $5.9\%$~(CUB). Furthermore, \texttt{VGSE-SMO} (\texttt{Attribute}) also outperform \texttt{Attribute} on both AWA2 and CUB dataset, i.e., we achieve $66.7\%$~(ZSL) on AWA2, compared to human attributes with $62.8\%$. 
The results demonstrate that our \texttt{VGSE-SMO} embeddings coupled with visually-grounded information can not only outperform the unsupervised word embeddings, but also improve over human attributes in transferring knowledge under the zero-shot setting.

\subsection{Qualitative Results}
\label{sec:quali}

\begin{figure}[t!]
  \centering
  \includegraphics[width=.8\linewidth]{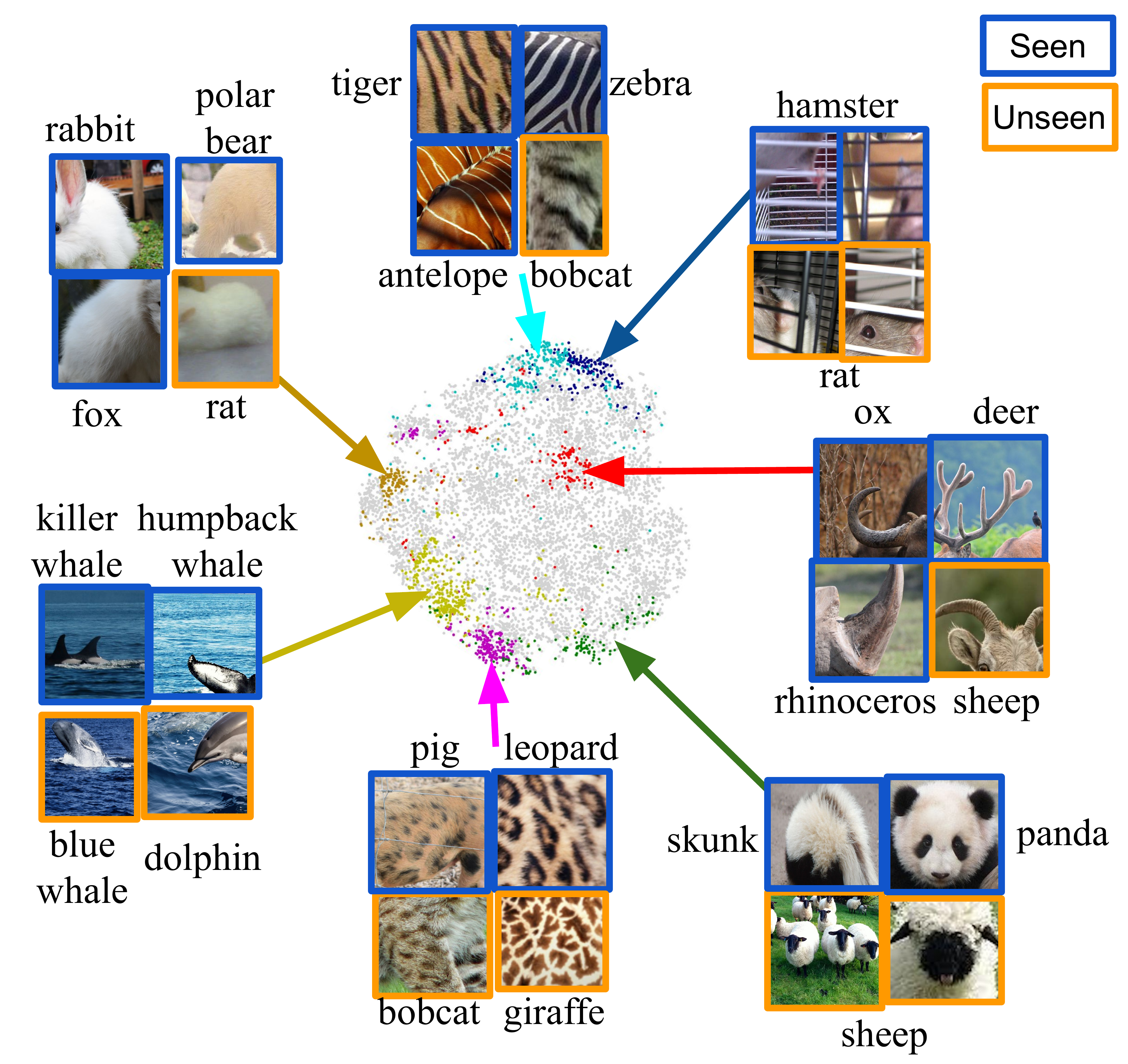}
    \caption{T-SNE embeddings of image patches from AWA2. Each colored dot region represents one visual cluster learnt by our \texttt{VGSE} model. We sample the seen~(in blue) and unseen images~(in orange) from the cluster center with their class names shown nearby. 
    } 
  \label{fig:tsne}
\end{figure}

In Figure~\ref{fig:tsne}, we show the 2D visualization of image patches in the AWA2, where $10,000$ image patches are presented by projecting their embeddings $a_{nt}$ onto two dimensions with t-SNE~\cite{tsne}. To picture their distribution on the embedding space, we sample several visual clusters~(dots marked in the same color) and the image patches from the cluster center of both seen and unseen categories.
Note that the unseen patches are not used to predict the unseen semantic embeddings, but only used for visualization here.

We observe that samples in the same cluster tend to gather together, indicating that the embeddings provide discriminative information. Besides, images patches in one cluster do convey consistent visual properties, though coming from disjoint categories. For instance, the \textit{white fur} appears on \textit{rabbit}, \textit{polar bear}, and \textit{fox} are clustered into one group, and the \textit{striped fur} from \textit{tiger}, \textit{zebra}, and \textit{bobcat} gather together because of their similar texture. 
We further observe that nearly all clusters consist images from more than one categories. For instance, the \textit{horns} from seen classes \textit{ox}, \textit{deer}, \textit{rhinoceros}, and unseen class \textit{sheep}, that with slightly different shape but same semantic, are clustered together. Similar phenomenon can be observed on the  \textit{spotted fur} and \textit{animals in ocean} clusters. It indicates that the clusters we learned contain semantic properties shared across seen classes, and can be transferred to unseen classes. Another interesting observation is that our \texttt{VGSE} clusters discover visual properties that my be neglected by human-annotated attributes, e.g., the \textit{cage} appear for \textit{hamsters} and \textit{rat}, and the \textit{black and white fur} not only appear on \textit{gaint panda} but also on \textit{sheeps}.

\subsection{Human Evaluation}
\label{sec:human}

To evaluate if our VGSE conveys consistent visual and semantic properties, we randomly pick 50 clusters, each equipped with 30 images from the cluster center, and ask 5 postgraduate students without prior knowledge of ZSL to examine the clusters and answer the following three questions.
Q1: Do images in this cluster contain consistent visual property?
Q2: Do images in this cluster convey consistent semantic information?
Q3: Please name the semantics you observed from the clusters, if your answer to Q2 is true. 
We do the same user study to 50 randomly picked clusters from the k-means clustering model. The results reveal that in $88.5\%$ and $87.0\%$ cases, users think our clusters convey consistent visual and semantic information. While for k-means clusters, the results are $71.5\%$ and $71.0\%$, respectively. The user evaluation results agree with the quantitative results in Table~\ref{tab:clustering_ablation}, which demonstrates that the class embeddings containing consistent visual and semantic information can significantly benefit the ZSL performance. Interestingly, by viewing \texttt{VGSE} clusters, users can easily discover semantics and even fine-grained attributes not depicted by human-annotated attributes, i.e., the \textit{fangs} and \textit{horns} in figure~\ref{fig:teaser}. Note that the whole process, i.e., naming 50 attributes for 40 classes, took less than 1 hour for each user. 

\section{Conclusion}
We develop a Visually-Grounded Semantic Embedding Network (VGSE) to learn distinguishing semantic embeddings for zero-shot learning with minimal human supervision. By clustering image patches with respect to their visual similarity, our network explores various semantic clusters shared between classes. Experiments on three benchmark datasets demonstrate that our semantic embeddings predicted from the class-relation module are generalizable to unseen classes, i.e., achieving significant improvement compared with word embeddings when trained with five models in both ZSL and GZSL settings. We further show that the visually augmented semantic embedding outperforms other semantic embeddings learned with minimal human supervision. The qualitative results verify that we discover visually consistent clusters that generalize from seen to unseen classes and can unearth the fine-grained properties not depicted by humans.


\section*{Acknowledgements}
This work has been partially funded by the ERC 853489 - DEXIM and by the DFG – EXC number 2064/1 – Project number 390727645.

{\small
\bibliographystyle{ieee_fullname}
\bibliography{egbib}
}

\end{document}